\def\year{2017}\relax
\documentclass[letterpaper]{article}
\usepackage{aaai17}
\usepackage{times}
\usepackage{centernot}
\usepackage{helvet}
\usepackage{amssymb}
\usepackage{courier}
\usepackage{amsmath}
\usepackage{bbm}
\DeclareMathOperator*{\argmax}{arg\,max}
\DeclareMathOperator*{\std}{std}
\DeclareMathOperator*{\argmin}{arg\,min}
\DeclareMathAlphabet{\mathpzc}{OT1}{pzc}{m}{it}
\newcommand{\fhatstar}{\hat{\rule{0ex}{1.7ex}\mkern-4mu f} {\vphantom{f}}^{*}}

\usepackage[]{tikz}
\usetikzlibrary{arrows}
\usetikzlibrary{positioning}
\usepackage{graphicx}
\usepackage{color}

\begin{document}
%
\title{Label-Free Supervision of Neural Networks with \\ Physics and Domain Knowledge}

\author{Russell Stewart , Stefano Ermon\\ 
Department of Computer Science, Stanford University\\
\{stewartr, ermon\}@cs.stanford.edu\\
}

\maketitle
\begin{abstract}
In many machine learning applications, labeled data is scarce and obtaining more labels is expensive.
We introduce a new approach to supervising neural networks by specifying constraints that should hold over the output space, rather than direct examples of input-output pairs. These constraints are derived from prior domain knowledge, e.g., from known laws of physics.
We demonstrate the effectiveness of this approach on real world and simulated computer vision tasks. We are able to train a convolutional neural network to detect and track objects without any labeled examples.
Our approach can significantly reduce the need for labeled training data, but introduces new challenges for encoding prior knowledge into appropriate loss functions.
\end{abstract}

\section{Introduction}
Applications of machine learning are often encumbered by the need for large amounts of labeled training data. Neural networks have made large amounts of labeled data even more crucial to success~\cite{krizhevsky2012imagenet,lecun2015deep}. 
Nonetheless, we observe that humans are often able to learn without direct examples, opting instead for high level instructions for how a task should be performed, or what it will look like when completed.
In this work, we ask whether a similar principle can be applied to teaching machines; can we supervise networks without individual examples by instead describing only the structure of desired outputs?



Contemporary methods for learning without labels often fall under the category of unsupervised learning. Autoencoders, for example, aim to uncover hidden structure in the data without having access to any label. Such systems succeed in producing 
highly compressed, yet informative representations of the inputs~\cite{kingma2013auto,le2013building}. However, these representations differ from ours as they are not explicitly constrained to have a particular meaning or semantics.

In this paper, we explicitly provide the semantics of the hidden variables we hope to discover, but still train without labels by learning from constraints (see \cite{shcherbatyi2016convexification} for an introduction to this idea). Intuitively, algebraic and logical constraints are used to encode structures and relationships that are known to hold because of prior domain knowledge.
The process of providing these necessary constraints may still require large amounts of domain specific engineering.

Nevertheless, by training without direct examples of the values our hidden (output) variables take, we gain several advantages over traditional supervised learning, including 1) a reduction in the amount of work spent labeling, and 2) an increase in generality, as a single set of 
constraints can be applied to multiple data sets without relabeling.
The primary contribution of this work is to demonstrate how constraint learning may be used to supervise neural networks across three practical computer vision tasks. We explore the challenge of simultaneously learning feature representations over raw data and avoiding trivial, low entropy solutions in the constraint space.  


\subsection{Problem Setup}
\begin{figure}
    \centering
    \includegraphics[width=0.4\textwidth]{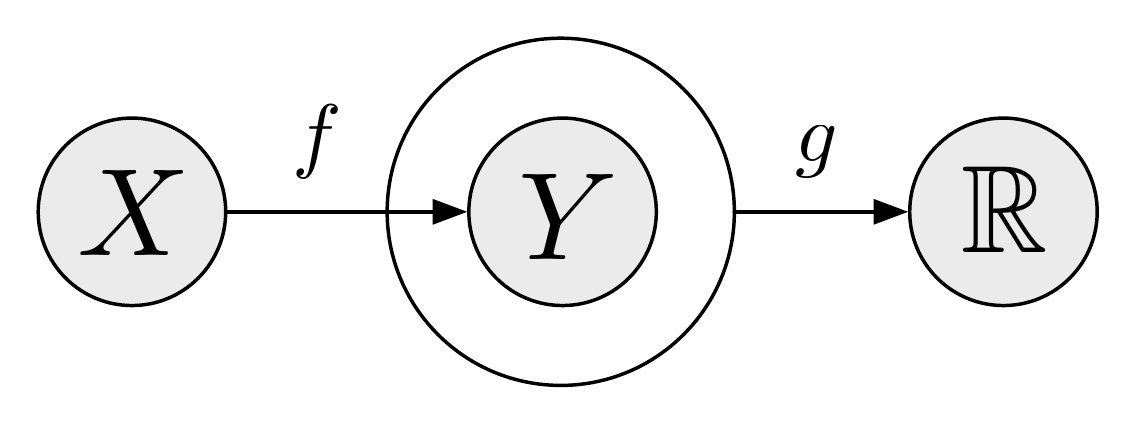}
    \caption{Constraint learning aims to recover the transformation $f$ without providing labels $y$. Instead, we look for a mapping $f$ that captures the structure required by  $g$.}
    \label{fig:label_free}
\end{figure}


In a traditional supervised learning setting, we are given a training set $D=\{(x_1, y_1), \cdots, (x_n, y_n)\}$ 
of $n$ training examples. Each example is a pair $(x_i,y_i)$ formed by an instance $x_i \in X$ and the corresponding output (label) $y_i \in Y$. The goal is to learn a function $f: X \rightarrow Y$ mapping inputs to outputs. To quantify performance, a loss function $\ell:Y \times Y \rightarrow \mathbb{R}$ is provided, and a mapping is found via
\begin{equation}\label{eq:fstar}
f^* = \argmin_{f \in \mathcal{F}} \sum_{i=1}^n \ell(f(x_i),y_i)
\end{equation}
where the optimization is over a pre-defined class of functions $\mathcal{F}$ (hypothesis class). In our case, $\mathcal{F}$ will be (convolutional) neural networks parameterized by their weights. The loss could be for example $\ell(f(x_i),y_i) = 1[f(x_i) \neq y_i]$.

By restricting the space of possible functions specifying the hypothesis class $\mathcal{F}$, we are leveraging prior knowledge about the specific problem we are trying to solve. Informally, the so-called No Free Lunch Theorems state that every machine learning algorithm must make such assumptions in order to work~\cite{wolpert2002supervised}. Another common way in which a modeler incorporates prior knowledge is by specifying an a-priori preference for certain functions in $\mathcal{F}$, incorporating a regularization term $R:\mathcal{F} \rightarrow \mathbb{R}$, and solving for
$
f^* = \argmin_{f \in \mathcal{F}} \sum_{i=1}^n \ell(f(x_i),y_i)  + R(f)
$. Typically, the regularization term $R:\mathcal{F} \rightarrow \mathbb{R}$ specifies a preference for ``simpler'' functions (Occam's razor). Regularization techniques such as dropout \cite{srivastava2014dropout} 
are key to avoid overfitting when training functions with a massive number of parameters.

%

In many ML settings, the input space $X$ is complex (images), while the output space $Y$ is simple (e.g., a binary classification problem where $Y=\{0,1\}$). Here, we are interested in structured prediction problems, where both $X$ and $Y$ are complex. 
For example, in our first experiment, $X$ corresponds to image sequences (video) and $Y$ to the height of an object as it is moving through the air. The goal is to identify a function $f^*$ that correctly maps frames to the corresponding height of the object. Clearly, the heights in each frame are not independent, and the sequence demonstrates a well-defined (algebraic) structure.
In fact, we known from elementary physics that any correct sequence of outputs forms a parabola. In principle, one could incorporate this prior knowledge by considering only valid subsets of sequences $Y^{\textrm{parabola}} \subset Y$, and restricting the hypothesis class $\mathcal{F}$ accordingly. This approach, however, would still require labels, and is difficult to combine with established and very successful regularization methods for neural networks.
 
In this paper, we model prior knowledge on the structure of the outputs by providing a weighted constraint function $g:X \times Y \rightarrow \mathbb{R}$, used to penalize ``structures'' that are not consistent with our prior knowledge. When $Y$ is a (multidimensional) discrete space (e.g., describing many potential binary attributes of an image) as in our third application, $g$ can be specified compactly using a graphical model approach, as a sum of weighted potential or constraints that only depend on a small subsets of the variables \cite{Richardson2006} \cite{lafferty2001conditional}.

The question we explore in this paper is whether this weak form of supervision is sufficient to learn interesting functions. While one clearly needs labels, $y$, to evaluate $f^*$, labels may not be necessary to discover $f^*$. If prior knowledge informs us that outputs of $f^*$ have other unique properties among functions in $\mathcal{F}$, we may use these properties for training rather than direct examples $y$. Specifically, we consider an unsupervised approach where the labels $y_i$ are not provided to us, and
optimize for a necessary property of the output, $g$ instead. That is, we search for
\begin{equation}\label{eq:fhatstar}
 \fhatstar = \argmin_{f \in \mathcal{F}} \sum_{i=1}^n g(x_i,f(x_i))+ R(f)
\end{equation}
If we are fortunate, $f \in \mathcal{F}$ optimizing (\ref{eq:fhatstar}) but not (\ref{eq:fstar}), will be either nonexistent or too strange to converge to in practice when combined with commonly used hypothesis classes (convolutional layers encoding translation invariance) and regularization terms $R$. When this is the case, we can use Stochastic Gradient Descent (SGD) to optimize (\ref{eq:fhatstar}) in place of (\ref{eq:fstar}), freeing us from the need for labels.

When optimizing (\ref{eq:fhatstar}) is not sufficient to find $f^*$, we will add additional regularization terms
%
%
to supervise the machine towards correct convergence. 
For example, as we will see in our person detection experiment, if $g$ is undesirably satisfied by a $f \equiv C$ constant output, we can add a term to favor outputs with higher entropy.
The process of designing  the loss $g$ and the regularization term $R$ is a form of supervision, and can require a significant time investment. But unlike hand labeling, it does not increase proportional to the size of the data, $|D|$, and can be applied to new data sets often without modification.

\begin{figure*}
    \centering
    \includegraphics[width=0.8\textwidth]{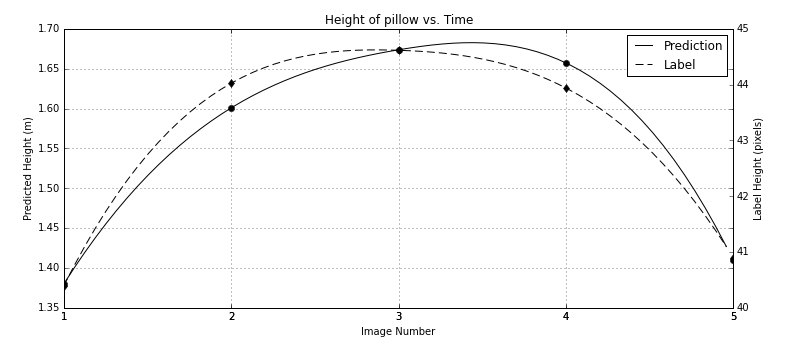}
    \includegraphics[width=0.9\textwidth]{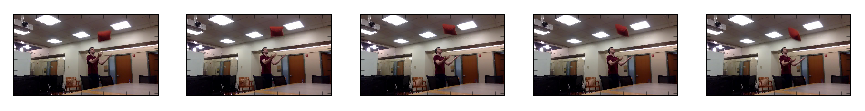}
    \caption{As the pillow is tossed, the height forms a parabola over time. We exploit this structure to independently predict the pillow's height in each frame without providing labels.} 
    \label{fig:cushion}
\end{figure*}

\section{Experiments}
The goal of our method is to train a network, $f$, mapping from inputs to outputs that we care about, without needing direct examples of those outputs. In our first two experiments, we construct a mapping from an image to the location of an object it contains. Learning is made possible by exploiting structure that holds in images over time. 
In our third experiment, we map an image to two boolean variables describing whether or not the image contains two special objects. Learning exploits the unique causal semantics existing between these objects. Across our experiments, we provide labels {\it only} for the purpose of evaluation.

\subsection{Tracking an object in free fall}
\begin{figure*}[ht]
    \centering
    \includegraphics[width=0.8\textwidth]{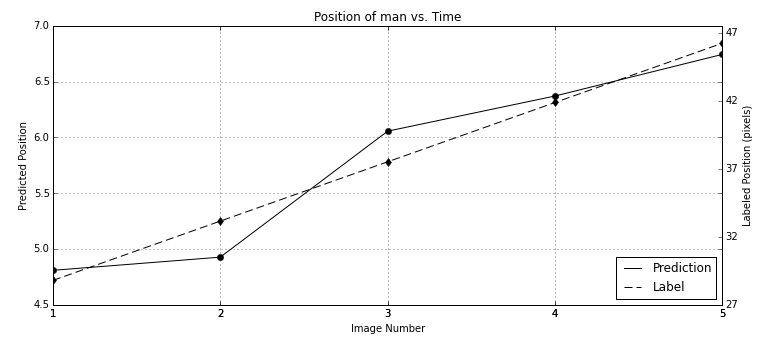}
    \includegraphics[width=0.9\textwidth]{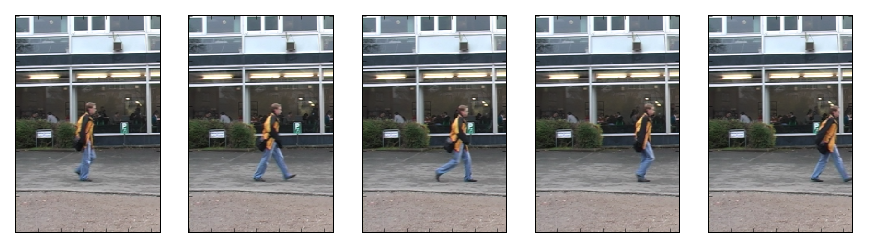}
    \caption{The network independently predicts the position of the walking man in each frame. Supervision tells the network that outputs must describe an object moving at constant (but non-zero) velocity.}
    \label{fig:man}
\end{figure*}
In our first experiment, we record videos of an object being thrown across the field of view, and aim to learn the object's height in each frame. Our goal is to obtain a regression network mapping from $R^{\text{height} \times \text{width} \times 3} \rightarrow \mathbbm{R}$, where $\text{height}$ and $\text{width}$ are the number of vertical and horizontal pixels per frame, and each pixel has 3 color channels. We will train this network as a structured prediction problem operating on a sequence of $N$ images to produce a sequence of $N$ heights, 
$\left(R^{\text{height} \times \text{width} \times 3} \right)^N \rightarrow \mathbbm{R}^N$, and each piece of data $x_i$ will be a vector of images, $\mathbf{x}$.
Rather than supervising our network with direct labels, $\mathbf{y} \in \mathbb{R}^N$, we instead supervise the network to find an object obeying the elementary physics of free falling objects. Because gravity acts equally on all objects, we need not encode the object's mass or volume.

An object acting under gravity will have a fixed acceleration of $a = -9.8 m / s^2$, and the plot of the object's height over time will form a parabola:

$$\mathbf{y}_i = y_0 + v_0(i\Delta t) + a(i\Delta t)^2$$

$\\$
\noindent This equation provides a necessary constraint, which the correct mapping $f^*$ must satisfy.
We thus train $f$ by making incremental improvements in the direction of better satisfying this equation. 


Given any trajectory of $N$ height predictions, $f(\mathbf{x})$, we fit a parabola with fixed curvature to those predictions, and minimize the resulting residual. Formally, we specify $\mathbf{a} = [a\Delta t^2, a(2 \Delta t)^2, \ldots, a(N \Delta t)^2]$ 
and set 

\begin{equation}\label{eq:linear}
\mathbf{\hat{y}} = \mathbf{a} + \mathbf{A} (\mathbf{A}^T\mathbf{A})^{-1} \mathbf{A}^T (f(\mathbf{x}) - \mathbf{a})
\end{equation}

\noindent where
$$\mathbf{A} = 
\left[ {\begin{array}{*{20}c}
   \Delta t & 1  \\
   2\Delta t & 1  \\
   3\Delta t & 1  \\
   \vdots & \vdots \\
   N\Delta t & 1  \\

 \end{array} } \right]
$$

\noindent The constraint loss is then defined as

$$g(\mathbf{x},f(\mathbf{x})) = g(f(\mathbf{x})) = \sum_{i=1}^{N} |\mathbf{\hat{y}}_i - f(\mathbf{x})_i|$$
where we note that the vector $\mathbf{\hat{y}}$ from (\ref{eq:linear}) is a function of the predictions $f(\mathbf{x})$, rather than ground truth labels.
\noindent Because $g$ is differentiable almost everywhere, we can optimize  equation~(\ref{eq:fhatstar}) with SGD. 
Surprisingly, we find that when combined with existing regularization methods for neural networks, this optimization is sufficient to recover $f^*$ up to an additive constant $C$ (specifying what object height corresponds to 0). Qualitative results from our network applied to fresh images after training are shown in Fig.~(\ref{fig:cushion})

\subsubsection{Training details}
Our data set is collected on a laptop webcam running at 10 frames per second ($\Delta t = 0.1s$). We fix the camera position and record 65 diverse trajectories of the object in flight, totalling 602 images. For each trajectory, we train on randomly selected intervals of $N=5$ contiguous frames. Our data will be made public upon publication.

Images are resized to $56 \times 56$ pixels before going into a small, randomly initialized neural network with no pretraining.
We use 3 Conv/ReLU/MaxPool blocks followed by 2 Fully Connected/ReLU layers with probability 0.5 dropout and a single regression output.
We group trajectories into batches of size 16, for a total of 80 images on each iteration of training. We use the Adam optimizer \cite{kingma2014adam} in TensorFlow \cite{abadi2016tensorflow} with a learning rate of 0.0001 and train for 4,000 iterations.

\subsubsection{Evaluation}
For evaluation, we manually labeled the height of our falling objects in pixel space. Note that labeling the true height in meters requires knowing the object's distance from the camera, so we instead evaluate by measuring the correlation of predicted heights with ground truth pixel measurements.
All results are evaluated on test images not seen during training.
Note that a uniform random output would have an expected correlation of 12.1\%.
Our network results in a correlation of 90.1\%.
For comparison, we also trained a supervised network on the labels to directly predict the height of the object in pixels. This network achieved a correlation of 94.5\%, although this task is somewhat easier as it does not require the network to compensate for the object's distance from the camera.

This experiment demonstrates that one can teach a neural network to extract object information from real images by writing down only the equations of physics that the object obeys.


\subsection{Tracking the position of a walking man}
In our second experiment, we now seek to extend the detection of free falling objects to other types of motion. 
We will aim to detect the horizontal position of a person walking across a frame without providing direct labels $y \in \mathbb{R}$. To this end, we exploit structure that holds over time by assuming the person will be walking at a constant velocity over short periods of time.
We thus formulate a structured prediction problem $f: \left(R^{\text{height} \times \text{width} \times 3} \right)^N \rightarrow \mathbbm{R}^N$, and treat each training instances $x_i$ as a vector of images, $\mathbf{x}$, being mapped to a sequence of predictions, $\mathbf{y}$. 

We work with a previously collected data set where we observed that the constant velocity assumption approximately holds. Given the similarities to our first experiment with free falling objects, we might hope to simply remove the gravity term from equation (\ref{eq:linear}) and retrain. 
However, in this case, that is not possible, as the constraint provides a necessary, but not sufficient, condition for convergence.

Given any sequence of correct outputs, $(\mathbf{y}_1, \ldots, \mathbf{y}_N)$, the modified sequence, $(\lambda * \mathbf{y}_1 + C, \ldots, \lambda * \mathbf{y}_N + C)$ ($\lambda, C \in \mathbbm{R}$) will also satisfy the constant velocity constraint.
In the worst case, when $\lambda = 0$, $f \equiv C$, and the network can satisfy the constraint while having no dependence on the image. Empirically, we observe that $f \equiv C$ is very easy to learn, and if we do not explicitly guard against this trivial solution, the network will always converge to it. 

We encode the desire for a nontrivial output by adding two additional loss terms. First, we reward the network for outputting a greater standard deviation of values across the sequence:

\begin{equation*}
\begin{split}
h_1(\mathbf{x}) = -\text{std}(f(\mathbf{x}))
\end{split}
\end{equation*}

\noindent However, this objective introduces a problem by providing infinite reward as $\lambda \rightarrow \infty$. We counterbalance this effect by requiring that the output across the image sequence to lie within a fixed range, $[0, 10]$:

\begin{equation*}
\begin{split}
h_2(\mathbf{x}) = \hphantom{'} & \text{max}(\text{ReLU}(f(\mathbf{x}) - 10)) \hphantom{\text{ }}+ \\
       & \text{max}(\text{ReLU}(0 - f(\mathbf{x})))
\end{split}
\end{equation*}

\noindent The final loss is thus:

\begin{equation*}
\begin{split}
g(\mathbf{x}) = \hphantom{'} & ||(\mathbf{A} (\mathbf{A}^T\mathbf{A})^{-1} \mathbf{A}^T - \mathbf{I}) * f(\mathbf{x})||_1 \hphantom{\text{ }}+ \\
       &  \gamma_1 * h_1(\mathbf{x}) 
       \hphantom{\text{ }}+ \\
       &  \gamma_2 * h_2(\mathbf{x})
\end{split}
\end{equation*}

\noindent We alternatively might have measured the constraint loss in a scale-invariant manner (e.g. by whitening outputs before measuring the inertial loss). This is consistent with the principle that there are multiple options for sufficiency terms to guide convergence.
\subsubsection{Training Details}
As shown in Fig~(\ref{fig:man}), our network is indeed able to discover the horizontal position of person walking in front of the camera.
Our data set contains 11 trajectories across 6 distinct scenes, totalling 507 images resized to $56 \times 56$. We train our network to output linearly consistent positions on 5 strided frames from the first half of each trajectory, and evaluate on the second half. 
We set the boundary violation penalty, $\gamma_2 = 0.8$, to be greater than the standard deviation bonus, $\gamma_1 = 0.6$, leading the network to find the solution with maximal $\lambda$ not violating the boundary constraint.
We choose exactly the same hyperparameters (dropout ratio, number of iterations, number of hidden units, etc.) on both this experiment and the free fall experiment, demonstrating some degree of robustness to these parameters. 

\subsubsection{Evaluation}
Our test labels are measured in pixels, whereas our predictions are in arbitrary units up to affine transformation. Thus, we find the best affine transformation $(\alpha, \beta)$ mapping our predictions onto pixel space for each trajectory, and measure the correlation. Note that $\alpha$ and $\beta$ can differ between scenes, and thus this metric does not demonstrate a complete solution to the object detection problem.
Nonetheless, we find that our predictions are 95.4\% correlated with the ground truth. Surprisingly, the same network trained with direct supervision struggled more with generalization, and scored a correlation of 80.5\% on the test set (99.8\% on training). 
We attribute this decreased performance to overfitting on the small amount of training data available (11 trajectories), and would expect a near perfect correlation for a well trained supervised classifier.

This experiment demonstrates the possibility of learning to detect an inertial object without labels. Importantly, it also shows that even when the primary structural constraint is not sufficient to guide learning, we may impose additional terms to encourage a correct, nontrivial solution.

\subsection{Detecting objects with causal relationships}
In the previous experiments, we explored options for incorporating constraints pertaining to dynamics equations in real world phenomena, i.e., prior knowledge derived from elementary physics. 
Other sources of domain knowledge can in principle be used to provide supervision in the learning process. For example, significant efforts have been devoted in the past few decades to construct large knowledge bases \cite{lenat1995cyc,bollacker2008freebase}.
This knowledge is typically encoded using logical and constraint based formalisms.
Thus, in this third experiment, we explore the possibilities of learning from logical constraints imposed on single images. More specifically, we ask whether it is possible to learn from causal phenomena.


\begin{figure}[ht]
    \centering
    \includegraphics[width=1.0\columnwidth]{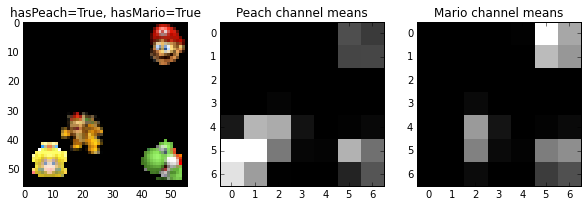}
    \includegraphics[width=1.0\columnwidth]{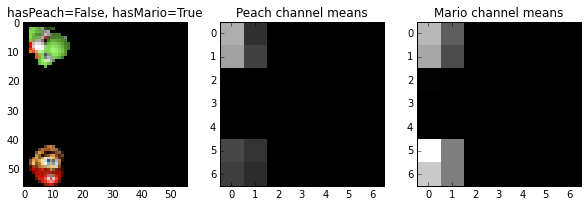}
    \caption{Whenever Peach (blond) shows up, Mario (red) comes around, but not vis versa. Yoshi (green) and Bowser (orange) appear randomly. The system trains with this high level knowledge and learns to answer whether each image contains Peach or Mario. The first column contains example images. The second and third columns show the attended locations for the Peach and Mario networks, respectively.}
    \label{fig:causal}
\end{figure}

We provide images containing a stochastic collection of up to four characters: Peach, Mario, Yoshi, and Bowser, with each character having small appearance changes across frames due to rotation and reflection. Example images can be seen in Fig. (\ref{fig:causal}).
While the existence of objects in each frame is non-deterministic, the generating distribution encodes the underlying phenomenon that Mario will always appear whenever Peach appears. Our aim is to create a pair of neural networks $f_1$, $f_2$ for identifying Peach and Mario, respectively. The networks, $f_k: R^{\text{height} \times \text{width} \times 3} \rightarrow \{0, 1\}$, map the image to the discrete boolean variables, $y_1$ and $y_2$.
Rather than supervising with direct labels, we train the networks by 
constraining their outputs to have the logical relationship $y_1 \Rightarrow y_2$.
This problem is challenging because the networks must simultaneously learn to recognize the characters and select them according to logical relationships. 

Merely satisfying the constraint $y_1 \Rightarrow y_2$ is not sufficient to certify learning. For example, the system might falsely report the constant output, $y_1 \equiv 1, y_2 \equiv 1$ on every image. Such a solution would satisfy the constraint, but say nothing about the presence of characters in the image.

To avoid such trivial solutions, we add three loss terms: $h_1$, $h_2$, and $h_3$.
$h_1$ forces rotational independence of the output by applying a random horizontal and vertical reflection $\rho$, to images. This encourages the network to focus on existence of objects, rather than location.
$h_2$ and $h_3$ allows us to avoid trivial solutions by encouraging high standard deviation and high entropy outputs, respectively. Given a batch of $M = 16$ images which we denote $\mathbf{x}$, we define

\begin{equation*}
\begin{split}
h_1(\mathbf{x}, k) &= \frac{1}{M}\sum_i^M |Pr[f_k(\mathbf{x}) = 1] - Pr[f_k(\rho(\mathbf{x})) = 1]| \\
h_2(\mathbf{x}, k) &= -\std_{i \in [1 \dots M]}(Pr[f_k(\mathbf{x}_i) = 1]) \\
h_3(\mathbf{x}, v) &= \frac{1}{M}\sum_i^{M} (Pr[f(\mathbf{x}_i) = v] - \frac{1}{3} + (\frac{1}{3} - \mu_v))^2 \\
\mu_{v} &= \frac{1}{M}\sum_i^{M} \mathbbm{1}\{v = \argmax_{v' \in \{0, 1\}^2} Pr[f(\mathbf{x}) = v']\} \\
\end{split}
\end{equation*}



%

%

\noindent After applying these constraints, one problem remains. The constraints are invariant to logical permutations (e.g. given a correct solution, $y_1^*, y_2^*$, the false solution $\hat{y_1} = y_1^*, \hat{y_2} = (y_1^* \wedge y_2^*) \vee (\neg y_1^* \wedge \neg y_2^*)$ would satisfy the equations equally well). We address this by forcing each boolean variable to derive it's value from a single region of the image
(each character can be identified from a small region in the image.)
The Peach network, $f_1$, runs a series of convolution and pooling layers to reduce the original input image to a $7 \times 7 \times 64$ grid.
We find the $64$-dimensional spatial vector with the greatest mean and use the information contained in it to predict the first binary variable. Examples of channel means for the Mario and Peach networks can be seen in Fig.~(\ref{fig:causal}). The Mario network $f_2$ performs the same process. But if the Peach networks claims to have found an object, $f_2$ is prevented from picking any vector within 2 spaces of the location used by the first vector.


%

The final loss function is given by:

\begin{equation*}
\begin{split}
g(\mathbf{x}) & = \mathbbm{1}\{f_1(\mathbf{x}) \centernot\implies f_2(\mathbf{x})\} \hphantom{\text{ }} + \\
& \sum_{k \in \{1, 2\}} \gamma_1 h_1(\mathbf{x}, k) + \gamma_2 h_2(\mathbf{x}, k) + 
\hspace{-0.7em} \sum_{v \neq \{1,0\}} \hspace{-0.7em} \gamma_3 * h_3(\mathbf{x}, v)
\end{split}
\end{equation*}

\noindent We construct both $f_1$ and $f_2$ as neural networks with 3 Conv/ReLU/MaxPool blocks as in our first two experiments. These blocks are followed by 2 Fully Connected/ReLU units, although the first fully connected layer receives input from only one spatial vector as described above.

\subsubsection{Evaluation}
Our input images, shown in Fig.~(\ref{fig:causal}), are $56 \times 56$ pixels.
We set $\gamma_1 = 0.65, \gamma_2 = 0.65, \gamma_3 = 0.95$, and training converges after 4,000 iterations.
On a test set of 128 images, the network learns to map each image to a correct description of whether the image contains Peach and Mario.

This experiment demonstrates that networks can learn from constraints that operate over discrete sets with potentially complex logical rules. Removing constraints $h_1 \text{, } h_2 \text {, or } h_3$ will cause learning to fail. Thus, the experiment also shows that sophisticated sufficiency conditions can be key to success when learning from constraints.

\section{Related Work}

In this work, we have presented a new strategy for incorporating domain knowledge in three computer vision tasks.
The networks in our experiments learn without labels by exploiting high level instructions in the form of constraints.

Constraint learning is a generalization of supervised learning that allows for more creative methods of supervision.
For example, multiple-instance learning as proposed by \cite{dietterich1997solving,zhou2007relation} allows for more efficient labeling by providing annotations over groups of images and learning to predict properties that hold over at least one input in a group, rather than providing individual labels. In rank learning, labels may given as orderings between inputs with the objective being to find an embedding of inputs that respects the ordering relation \cite{joachims2002optimizing}.
Inductive logic programming approaches rely on logical formalisms and constraints to represent background knowledge and learn hypotheses from data~\cite{muggleton1994inductive,de2008logical,de2003probabilistic}.
Various types of constraints have also been used extensively to guide unsupervised learning algorithms, such as clustering and dimensionality reduction techniques~\cite{lee2001algorithms,basu2008constrained,zhi2013clustering,ermon2015pattern}.

Each of our experiments differs from such classical examples of constraint learning by jointly 1) leveraging the representation learning abilities of modern neural networks, and 2) adding sufficiency terms when the primary constraint is merely necessary. But the use of constraint learning for neural networks has also suggested by several other recent works.

\cite{Kotzias2015} used constraint learning to train deep networks in a natural language setting. Sentiment labels on reviews were used to analyze the sentiment of individual sentences comprising those reviews.
\cite{linlearning} and \cite{zhuang2016fast} trained deep convolutional neural networks to construct high level compressed embeddings of images without using labels.
\cite{linlearning} encoded constraints such as invariance of embeddings to image rotations, high entropy outputs, and high standard deviation outputs to learn these embeddings. Our experiments build on these ideas in a context where we can use prior knowledge such as physical dynamics to further constrain the output's semantics.

The Deep Q-Network (DQN) of \cite{dqn15} provides another inspirational example for training neural networks with constraints rather than direct labels.
The DQN may be described as an optimization of equation \ref{eq:fhatstar} by:

\begin{itemize}
\item{$X$: $(x_t, x_{t+1}) \in (\mathbbm{R}^{\text{height} \times \text{width} \times 3})^2 \text{ (a pair of sequential}\\
\hphantom{X\text{: }}\text{states)}$}
\item{$Y$: $(\mathbbm{R}^{|a|})^2$ (the expected future rewards from each state)}
\item{$f$: (convolutional) neural net with $|a|$ outputs}
\item{$g \text{: } f(x_t)_a - (\gamma * \argmax_{a'} f(x_{t+1})_{a'} + r(x_t))\\ \hphantom{g\text{: }}\text{(the Bellman equation)}$}
\end{itemize}

\noindent DQN's demonstrate that by imposing the right constraint $g$, one can transform weak labels of the form $r(x_t)$ into a rich planning algorithm over raw images.

Thus, a growing volume of work proposes the use of nontraditional loss functions for neural networks. The strategies outlined in the diverse set of references above each fall under the generic method of constraint learning. Our experiments encourage an even broader range of future applications where the primary constraint is necessary, but not sufficient for learning.

\section{Conclusion}
We have introduced a new method for using physics and other domain constraints to supervise neural networks.
Future challenges include extending these results to larger data sets with multiple objects per image, and simplifying the process of picking sufficiency terms for new and interesting problems.
By freeing the operator from collecting labels, our small scale experiments show promise for the future of training neural networks with weak supervision.

\section{Acknowledgments}
This work was supported by a grant from the SAIL-Toyota Center for AI Research. The authors would like to thank Aditya Grover and Tudor Achim for helpful discussions.




\bibliographystyle{aaai}
\bibliography{refs}

\end{document}